\def\cvprPaperID{7892}
\def\confName{CVPR}
\def\confYear{2023}

\def\authorBlock{
        Kun Su$^1$\thanks{Work done while interning at MIT-IBM Watson AI Lab} \qquad
    Kaizhi Qian$^2$ \qquad
    Eli Shlizerman$^1$ \qquad
    Antonio Torralba$^3$ \qquad
    Chuang Gan$^{2,4}$ \qquad \\
    $^1$University of Washington \qquad
    $^2$MIT-IBM Watson AI Lab \qquad
    $^3$MIT \qquad
    $^4$UMass Amherst\\
}

\newif\ifreview 
\newif\ifarxiv 
\newif\ifcamera \newcommand{\cameraready}{\cameratrue}
\newif\ifrebuttal 

\cameraready 

\pdfoutput=1
\documentclass[10pt,twocolumn,letterpaper]{article}
\ifreview \usepackage[review]{cvpr} \fi
\ifarxiv \usepackage[pagenumbers]{cvpr} \fi
\ifrebuttal \usepackage[rebuttal]{cvpr} \fi
\ifcamera \usepackage{cvpr} \fi

\usepackage{graphicx}
\usepackage{amsmath}
\usepackage{amssymb}
\usepackage{booktabs}

\usepackage{times}
\usepackage{microtype}
\usepackage{epsfig}
\usepackage[table,xcdraw]{xcolor}
\usepackage{caption}
\usepackage{float}
\usepackage{placeins}
\usepackage{color, colortbl}
\usepackage{stfloats}
\usepackage{enumitem}
\usepackage{tabularx}
\usepackage{xstring}
\usepackage{multirow}
\usepackage{xspace}
\usepackage{url}
\usepackage{subcaption}
\usepackage{xcolor}
\usepackage[hang,flushmargin]{footmisc}
\usepackage{amsmath}
\ifcamera \usepackage[accsupp]{axessibility} \fi

% ...

\ifarxiv  \fi

\newcommand{\R}[1]{{%
    \textbf{%
        \ifstrequal{#1}{1}{\textcolor{red}{R#1}}{%
        \ifstrequal{#1}{2}{\textcolor{blue}{R#1}}{%
        \ifstrequal{#1}{3}{\textcolor{magenta}{R#1}}{%
        \ifstrequal{#1}{4}{\textcolor{teal}{R#1}}{%
                           \textcolor{cyan}{R#1}%
        }}}}%
    }%
}}

\definecolor{mygray}{gray}{0.6}
\definecolor{mygray-bg}{gray}{0.9}

\usepackage{xr-hyper}

\makeatletter
\newcommand*{\addFileDependency}[1]{
  \typeout{(#1)}
  \@addtofilelist{#1}
  \IfFileExists{#1}{}{\typeout{No file #1.}}
}

\makeatother

\usepackage[pagebackref,breaklinks,colorlinks]{hyperref}
\usepackage[capitalize]{cleveref}
\crefname{section}{Sec.}{Secs.}
\crefname{table}{Table}{Tables}
\crefname{figure}{Fig.}{Figs.}

\begin{document}
\title{Physics-Driven Diffusion Models for Impact Sound Synthesis from Videos}
\author{\authorBlock}

\maketitle


\begin{abstract}
Modeling sounds emitted from physical object interactions is critical for immersive perceptual experiences in real and virtual worlds. Traditional methods of impact sound synthesis use physics simulation to obtain a set of physics parameters that could represent and synthesize the sound. However, they require fine details of both the object geometries and impact locations, which are rarely available in the real world and can not be applied to synthesize impact sounds from common videos. On the other hand, existing video-driven deep learning-based approaches could only capture the weak correspondence between visual content and impact sounds since they lack of physics knowledge. In this work, we propose a physics-driven diffusion model that can synthesize high-fidelity impact sound for a silent video clip. In addition to the video content, we propose to use additional physics priors to guide the impact sound synthesis procedure. The physics priors include both physics parameters that are directly estimated from noisy real-world impact sound examples without sophisticated setup and learned residual parameters that interpret the sound environment via neural networks. We further implement a novel diffusion model with specific training and inference strategies to combine physics priors and visual information for impact sound synthesis. Experimental results show that our model outperforms several existing systems in generating realistic impact sounds. Lastly, the physics-based representations are fully interpretable and transparent, thus allowing us to perform sound editing flexibly. We encourage the readers
visit our project page \footnote{\url{https://sukun1045.github.io/video-physics-sound-diffusion/}}
to watch demo videos with the audio turned on to experience the result.
\end{abstract}
\section{Introduction}
\label{sec:intro}
\begin{figure}
    \centering
    \includegraphics[width=\linewidth]{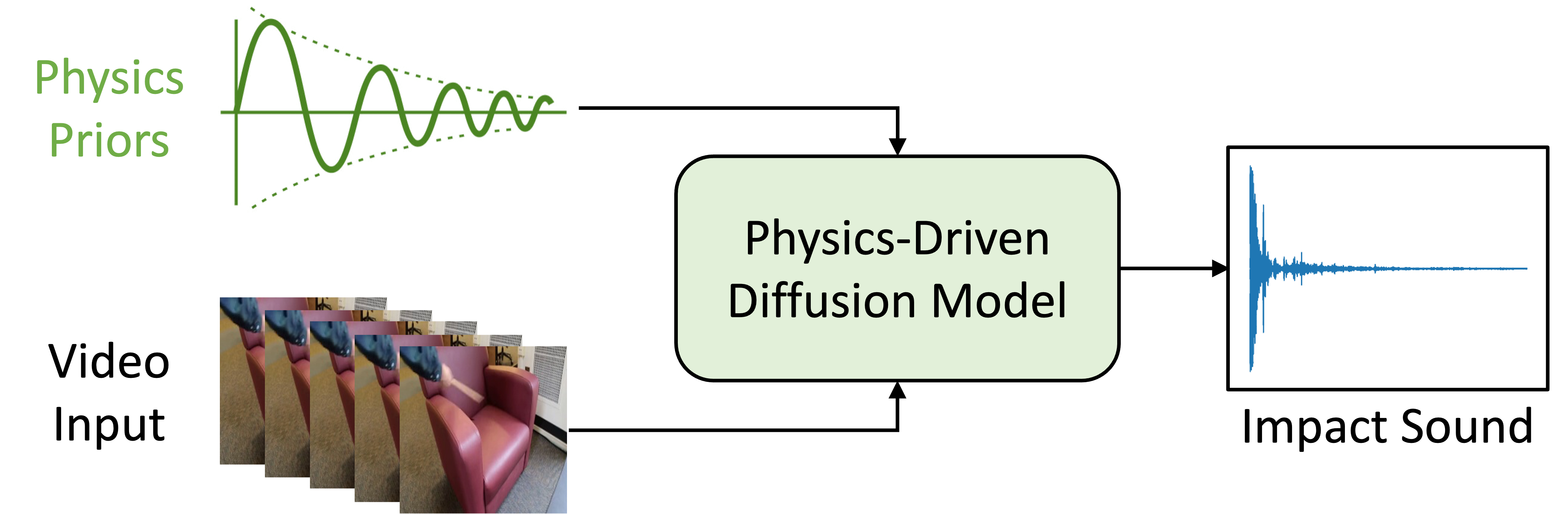}
    \caption{The physics-driven diffusion model takes physics priors and video input as conditions to synthesize high-fidelity impact sound. Please also see the supplementary video and materials with sample results.}
    \label{fig:teaser}
\end{figure}
Automatic sound effect production has become demanding for virtual reality, video games, animation, and movies. Traditional movie production heavily relies on talented Foley artists to record many sound samples in advance and manually perform laborious editing to fit the recorded sounds to visual content. Though we could obtain a satisfactory sound experience at the cinema, it is labor-intensive and challenging to scale up the sound effects generation of various complex physical interactions. 

Recently, much progress has been made in automatic sound synthesis, which can be divided into two main categories. The first category is physics-based modal synthesis methods~\cite{van2001foleyautomatic, o2001synthesizing, o2002synthesizing}, which are often used for simulating sounds triggered by various types of object interactions. Although the synthesized sounds can reflect the differences between various interactions and the geometry property of the objects, such approaches require a sophisticated designed environment to perform physics simulation and compute a set of physics parameters for sound synthesis. It is, therefore, impractical to scale up for a complicated scene because of the time-consuming parameter selection procedure. On the other hand, due to the availability of a significant amount of impact sound videos in the wild, training deep learning models for impact sound synthesis turns out to be a promising direction. Indeed, several works have shown promising results in various audio-visual applications~\cite{zhu2021deep}. Unfortunately, most existing video-driven neural sound synthesis methods~\cite{zhou2018visual, chen2020generating} apply end-to-end black box model training and lack of physics knowledge which plays a significant role in modeling impact sound because a minor change in the impact location could exert a significant difference in the sound generation process. As a result, these methods are prone to learning an average or smooth audio representation that contains artifacts, which usually leads to generating unfaithful sound.

In this work, we aim to address the problem of automatic impact sound synthesis from video input. The main challenge for the learning-based approach is the weak correspondence between visual and audio domains since the impact sounds are sensitive to the undergoing physics. Without further physics knowledge, generating high-fidelity impact sounds from videos alone is insufficient. Inspired by physics-based sound synthesis methods using a set of physics mode parameters to represent and re-synthesize impact sounds~\cite{ren2013example, traer2019perceptually, clarke2022diffimpact}, we design a physics prior that could contain sufficient physics information to serve as a conditional signal to guide the deep generative model synthesizes impact sounds from videos. However, since we could not perform physics simulation on raw video data to acquire precise physics parameters, we explored estimating and predicting physics priors from sounds in videos. We found that such physics priors significantly improve the quality of synthesized impact sounds. For deep generative models, recent successes in image generation such as DALL-E 2 and Imagen~\cite{saharia2022photorealistic} show that Denoising Diffusion Probabilistic Models (DDPM) outperform GANs in terms of fidelity and diversity, and its training process is usually with less instability and mode collapse issues. While the idea of the denoising process is naturally fitted with sound signals, it is unclear how video input and physics priors could jointly condition the DDPM and synthesize impact sounds.

To address all these challenges, we propose a novel system for impact sound synthesis from videos. The system includes two main stages. In the first stage, we encode physics knowledge of the sound using physics priors, including estimated physical parameters using signal processing techniques and learned residual parameters interpreting the sound environment via neural networks. In the second stage, we formulate and design a DDPM model conditioned on visual input and physics priors to generate a spectrogram of impact sounds. Since the physics priors are extracted from the audio samples, they become unavailable at the inference stage. To solve this problem, we propose a novel inference pipeline to use test video features to query a physics latent feature from the training set as guidance to synthesize impact sounds on unseen videos. Since the video input is unseen, we can still generate novel impact sounds from the diffusion model even if we reuse the training set's physics knowledge.
\noindent In summary, our main contributions to this work are:
\vspace{-2mm}
\begin{itemize}[align=right,itemindent=0em,labelsep=2pt,labelwidth=1em,leftmargin=*,itemsep=0em] 

\item We propose novel physics priors to provide physics knowledge to impact sound synthesis, including estimated physics parameters from raw audio and learned residual parameters approximating the sound environment.

\item We design a physics-driven diffusion model with different training and inference pipeline for impact sound synthesis from videos. To the best of our knowledge, we are the first work to synthesize impact sounds from videos using the diffusion model.

\item Our approach outperforms existing methods on both quantitative and qualitative metrics for impact sound synthesis. The transparent and interpretable properties of physics priors unlock the possibility of interesting sound editing applications such as controllable impact sound synthesis.
\end{itemize}

\section{Related Work}
\subsection{Sound Synthesis from Videos}
Sound synthesis has been an ongoing research theme with a long history in audio research. Traditional approaches mainly use linear modal synthesis to generate rigid body sounds~\cite{van2001foleyautomatic}. While such methods could produce sounds reflecting the properties of impact sound objects such as the geometry difference, it is often the case that the simulation and engineering tuning on the initial parameters for the virtual object materials in the modal analysis is time-consuming. Suppose we are under a more complicated scene with many different sounding materials; the traditional approach can quickly become prohibitively intractable~\cite{ren2013example}. In recent years, deep learning approaches have been developed for sound synthesis. Owens et al.~\cite{owens2016visually} investigated predicting the sound emitted by interacting in the wild objects using a wood drumstick. However, instead of directly using LSTM to generate sound, they first predict sound features and then performed an exemplar-based retrieval algorithm. Instead of performing retrieval, our work directly generates the impact sounds. DiffImpact~\cite{clarke2022diffimpact} proposed a fully differentiable model for rigid objects' impact sounds based on physical principles. In addition to impact sound, a conditional generative adversarial network is proposed for cross-modal generation on music performances collected in a lab environment by Chen et al.~\cite{chen2017deep}. Moreover, natural sounds are explored by Zhou et al.~\cite{zhou2018visual} who introduced a SampleRNN-based method to directly predict audio waveform from Youtube videos data but the number of sound categories is limited to ten. Next, several works attempt to generate aligned audio to input videos via a perceptual loss~\cite{chen2018visually} and information bottleneck~\cite{chen2020generating}. More recently, music generation from visual input has also achieved various attentions~\cite{su2020audeo, gan2020foley, su2021does}.

\subsection{Audio-visual learning}
In recent years, methods for multi-modality learning have shown significance in learning joint representation for downstream tasks~\cite{radford2021learning}, and unlocked novel cross-modal applications such as visual captioning~\cite{yu2022coca, liu2021aligning}, visual question answering (VQA)~\cite{yi2018neural, chen2021grounding}, vision language navigation~\cite{anderson2018vision}, spoken question answering (SQA)~\cite{you2021mrd, chen2021self}, healthcare AI~\cite{li2020behrt, liu2021auto, you2022mine, youclass, you2023rethinking}, etc. In this work, we are in the field of audio-visual learning, which deals with exploring and leveraging both audio and video correlation at the same time. For example, earlier work from Owens et al.~\cite{owens2016ambient} tried using clustered sound to learn visual representations from unlabeled video data, and similarly, Aytar et al.~\cite{aytar2016soundnet} leveraged the scene to learn the audio representations. Later, ~\cite{arandjelovic2017look} investigated audio-visual joint learning the visual by training a visual-audio correspondence task. More recently, several works have also explored sound source localization in images or videos in addition to the audio-visual representations~\cite{izadinia2012multimodal, hershey1999audio, senocak2018learning}. Such works include a lot of applications such as biometric matching~\cite{nagrani2018seeing}, visually-guided sound source separation~\cite{zhao2018sound, gan2020music, gao2018learning, xu2019recursive}, understanding physical scene via multi-modal~\cite{gan2020threedworld}, auditory vehicle tracking~\cite{gan2019self}, multi-modal action recognition~\cite{long2018attention, long2018multimodal, gao2020listen}, audio-visual event localization~\cite{tian2018audio}, audio-visual co-segmentation~\cite{rouditchenko2019self}, audio inpainting~\cite{zhou2019vision}, and audio-visual embodied navigation~\cite{gan2020look}.

\subsection{Diffusion Model}
The recently explored diffusion probabilistic models (DPMs)~\cite{sohl2015deep} have served as a powerful generative backbone that achieves promising results in various generative applications~\cite{kong2020diffwave, mittal2021symbolic, lee2021nu, ho2020denoising, nichol2021improved, dhariwal2021diffusion, ho2022cascaded}, outperforming GANs in terms of fidelity and diversity. More intriguing, the training process is usually with less instability and mode collapse issues. Compared to the unconditional case, conditional generation is usually applied in more concrete and practical cross-modality scenarios. Most existing DPM-based conditional synthesis works~\cite{gu2022vector, dhariwal2021diffusion} learn the connection between the conditioning and the generated data implicitly by adding a prior to the variational lower bound. The above methods mostly focus on image applications, while audio is usually different in its temporal dependencies. Recently, there are several works that have explored to apply diffusion models to text-to-speech (TTS) synthesis~\cite{jeong2021diff, huang2022prodiff}. Unlike the task of text-to-speech synthesis, which contains a strong correlation between phonemes and speech, the correspondences between impact sounds and videos are weak. Therefore it is non-trivial to directly apply a conditional diffusion model to impact sound synthesis from videos. In this work, we found that only video condition is insufficient to synthesize high-fidelity impact sounds and additionally apply physics priors significantly improve the results. Moreover, due to the difficulty of predicting physics priors from video, we propose different training and testing strategies that could benefit the information of physics priors but also synthesize new impact sounds from the video input.
\label{sec:related}

\section{Method}
\begin{figure*}[!t]
    \centering
    \includegraphics[width=0.85\textwidth]{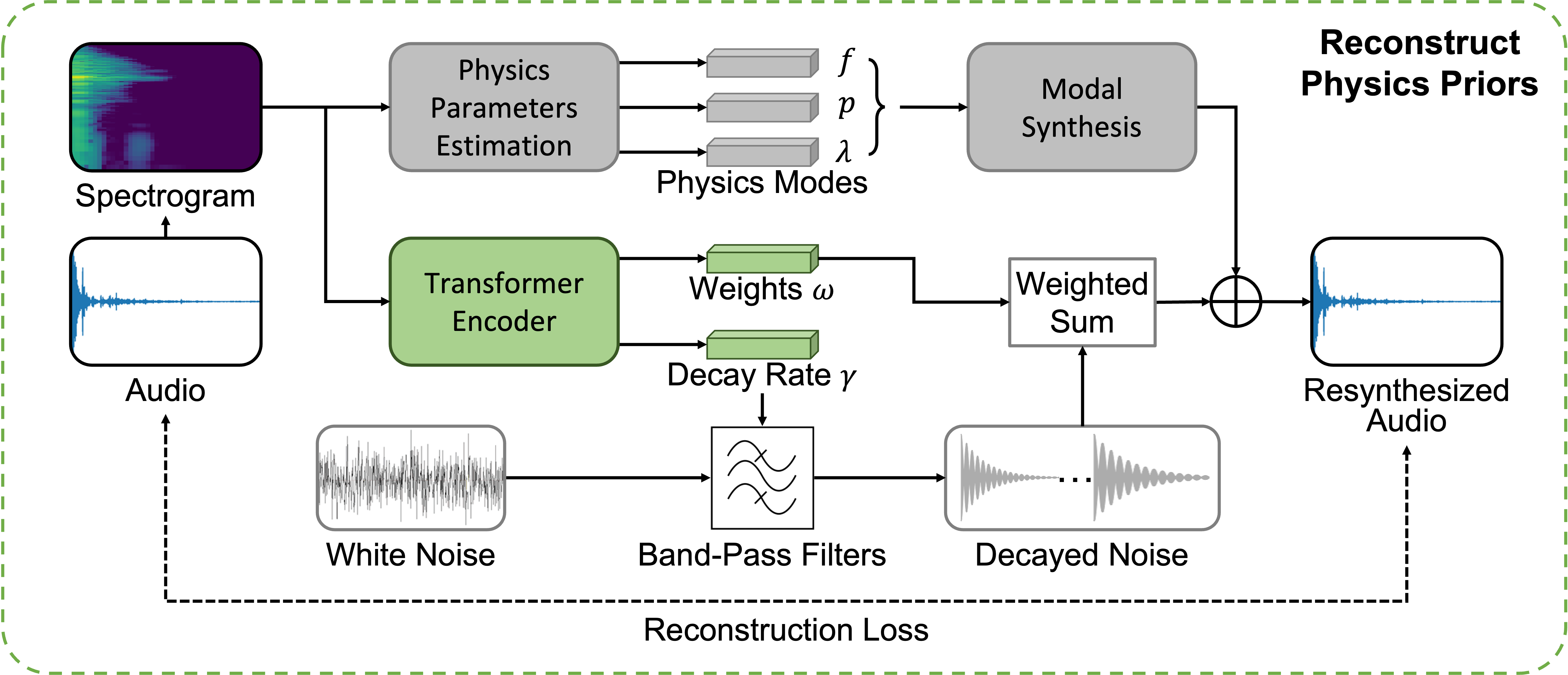}
    \caption{Reconstruction of physics priors by two components: $1$) We estimate a set of physics parameters (frequency, power, and decay rate) via signal processing techniques. $2$) We predict residual parameters representing the environment by a transformer encoder. A reconstruction loss is applied to optimize all trainable modules.}
    \label{fig:sound_physics}
\end{figure*}
Our method includes two main components: (a) physics priors reconstruction from sound (shown in Fig.~\ref{fig:sound_physics}), and (b) a physics-driven diffusion model for impact sound synthesis (shown in Fig.~\ref{fig:diffusion}). We first show how we can acquire physics priors from sounds In (a). Then in (b), we use reconstructed physics priors as additional information to the video input and guide the diffusion model to learn impact sound synthesis. Since no sound is available during the test time, we use different training and inference strategies to keep the benefit of physics priors and generate novel impact sounds.

\subsection{Reconstruct Physics Priors From Sound}

We aim to reconstruct physics from sound. There are two modules: $1$) the physics parameters estimation extracting modes parameters from audio waveform, and $2$) the residual parameters prediction learning to encode the environment information such as background noise and reverberation using neural networks.

\noindent\textbf{Physics Parameters Estimation}.
The standard linear modal synthesis technique is frequently used for modeling physics-based sound synthesis. The displacement $x$ in such a system can be computed with a linear equation described as follows:
\begin{align}
    M\Ddot{x} + C\Dot{x} + Kx = F, 
\end{align}
where $F$ represents the force, $M$ represents the mass, $C$ represents the damping, and $K$ represents the stiffness. With such a linear system, we can solve the generalized eigenvalue problem $KU = \Lambda MU$ and decouple it into the following form:
\begin{align}
    \Ddot{q} + (\alpha I + \beta \Lambda)\Dot{q} + \Lambda q = U^T F
\end{align}
where $\Lambda$ represents the diagonal matrix that contains eigenvalues of the system, $U$ represents the eigenvectors which can transform $x$ into the bases of decoupled deformation $q$ by matrix multiplication $x = Uq$.
\label{physics param estim}

After solving the decoupled system, we will obtain a set of modes that can be simply expressed as damped sinusoidal waves. The $i$-th mode can be expressed by:
\begin{align}
    q_i = p_i e^{-\lambda_i t} \sin (2\pi f_i t + \theta_i)
\end{align}
where $f_i$ is the frequency of the mode, $\lambda_i$ is the decaying rate, $p_i$ is the excited power, and $\theta_i$ is the initial phase. It is also common to represent $q_i$ under the decibel scale and we have
\begin{align}
    q_i = 10^{({p_i - \lambda_i t})/20} \sin (2\pi f_i t + \theta_i).
    \label{syn}
\end{align}
The frequency, power, and decay rate together define the physics parameter feature $\phi$ of mode $i$: $\phi = (f_i, p_i, \lambda_i)$ and we ignore $\theta_i$ since we assume the object is initially at rest and struck at $t=0$ and therefore it is usually treated as zero in the estimation process~\cite{ren2013example}.

Given a recorded audio waveform $s \in \mathbb{R}^{T}$, from which we first estimate physics parameters including a set of damped sinusoids with constant frequencies, powers, and decay rates. We first compute the log-spectrogram magnitude $S \in \mathbb{R}^{D \times N}$ of the audio by short-time-Fourier-transform (STFT), where $D$ is the number of frequency bins and $N$ is the number of frames. To capture sufficient physics parameters, we set the number of modes to be equal to the number of frequency bins. Within the range of each frequency bin, we identify the peak frequency $f$ from the fast Fourier transform (FFT) magnitude result of the whole audio segment. 
Next, we extract the magnitude at the first frame in the spectrogram to be the initial power $p$. Finally, we compute the decay time $\lambda$ for the mode according to the temporal bin when it reaches the silence ($-80$dB). At this point, we obtain $D$ modes physics parameters $\{(f_i, p_i, \lambda_i)\}^D_{i=1}$ and we can re-synthesize an audio waveform $\hat{s}$ using equation~\ref{syn}.

\noindent\textbf{Residual Parameters Prediction}. 
While the estimated modes capture most of the components of the impact sound generated by physical object interactions, the recorded audio in the wild has complicated residual components such as background noise and reverberation dependent on the sound environment which is critical for a real and immersive perceptual experience. Here we propose a learning-based approach to model such residual parameters. We approximate the sound environment component with exponentially decaying filtered noise similar to~\cite{clarke2022diffimpact}. We first randomly generate a Gaussian white noise $\mathcal{N}(0, 1)$ signal and perform a band-pass filter (BPF) to split it into $M$ bands. Then, for each band $m$, the residual component is formulated as
\begin{align}
    R_m = 10^{(-\gamma t)/20}\text{BPF}(\mathcal{N}(0, 1))_m
\end{align}
The accumulated residual components $R$ is a weighted sum of subband residual components
\begin{align}
    R = \sum^M_{m = 1} w_m R_m,
\end{align}
where $w_m$ is the weight coefficient of band $m$ residual component. Given the log-spectrogram $S\in \mathbb{R}^{D\times N}$ as input, we use a transformer-based encoder to encode each frame of the $S$. The output features are then averaged and two linear projections are used to estimate $\gamma \in \mathbb{R}^M$ and $w \in \mathbb{R}^M$. We minimize the error between $\hat{s}+R$ and $s$ by a multi-resolution STFT loss $L_{\text{mr-stft}}(\hat{s}+R, s)$ which has been shown effective in modeling audio signals in the time domain~\cite{yamamoto2020parallel}.
By estimating physics parameters and predicting residual parameters, we obtain the physics priors and it is ready to be a useful condition to guide the impact sound synthesis model to generate high-fidelity sounds from videos.

\subsection{Physics-Driven Diffusion Models}
\begin{figure*}[!t]
    \centering
    \includegraphics[width=\textwidth]{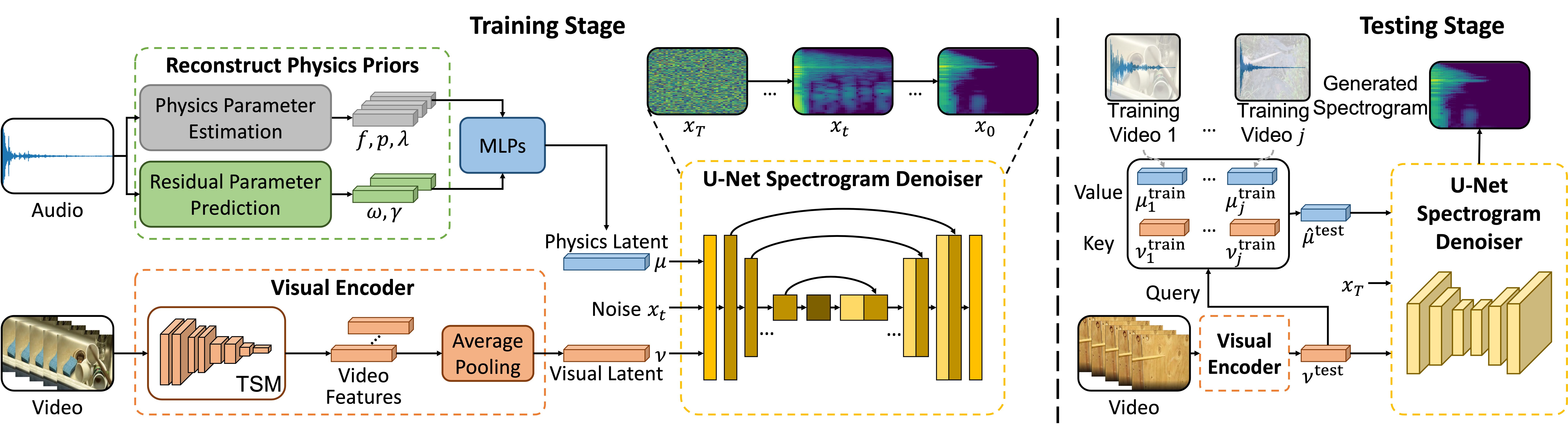}
    \caption{Overview of the physics-driven diffusion model for impact sound synthesis from videos. (left) During training, we reconstruct physics priors from audio samples and encode them into a physics latent. Besides, we use a visual encoder to extract visual latent from the video input. We apply these two latents as conditional inputs to the U-Net spectrogram denoiser. (right) During testing, we extract the visual latent from the test video and use it to query a physics latent from the key-value pairs of visual and physics latents in the training set. Finally, the physics and visual latents are used as conditional inputs to the denoiser and the denoiser iteratively generates the spectrogram.} 
    \label{fig:diffusion}
\end{figure*}
With the physics priors and video inputs, we propose a conditional Denoising Diffusion Probabilistic Model (DDPM) for impact sound synthesis. Our model performs a reverse diffusion process to guide the noise distribution to a spectrogram distribution corresponding to the input physics priors and video content. We encode all physics and residual parameters as a latent feature embedding with multi-layer perceptron (MLPs). The resulting physics latent vector is denoted by $\mu$. For video inputs, given a sequence of RGB frames, we use temporal-shift-module (TSM)~\cite{lin2019tsm} to efficiently extract visual features, which are then average pooled to compute a single visual latent representation $\nu$.

We show our physics-driven diffusion model for sound synthesis in Fig.~\ref{fig:diffusion}. The main component is a diffusion forward process that adds Gaussian noise $\mathcal{N}(0, I)$ at time steps $t=0, ..., T$ to a spectrogram $x$ with variance scale $\beta$. We can use a scheduler to change the variance scale at each time step to have $\beta_1, \beta_2, ..., \beta_T$~\cite{jeong2021diff}. We denote the spectrogram at diffusion time step $t$ as $x_t$. Given the spectrogram at time step $t-1$ as $x_{t-1}$, physics latent $\mu$, and visual latent $\nu$, the explicit diffusion process for spectrogram at time step $t$ can be written as $q(x_t | x_{t-1}, \mu, \nu)$. Since the complete diffusion process that takes $x_0$ to $x_T$ conditioned on $\mu$ and $\nu$ is a Markov process, we can factorize it into a sequence of multiplication $\prod_{t=1}^T q(x_t | x_{t-1})$.
To generate a spectrogram, we need the reverse process that aims to recover a spectrogram from Gaussian noise. The reverse process can be defined as the conditional distribution $p_{\theta} (x_{0:T-1} | x_T, \mu, \nu)$, and according to Markov chain property, it can be factorized into multiple transitions as follows:
\begin{align}
    p_{\theta} (x_0, ..., x_{T-1} | x_T, \mu, \nu) = \prod_{t=1}^T p_{\theta}(x_{t-1} | x_t , \mu, \nu).
\end{align}
Given the diffusion time-step with physics latent and visual latent conditions, a spectrogram is recovered from the latent variables by applying the reverse transitions $p_{\theta} (x_{t-1} | x_t, \mu, \nu)$. Considering the spectrogram distribution as $q(x_0 | \mu, \nu)$, we aim to maximize the log-likelihood of the spectrogram by learning a model distribution $p_{\theta} (x_0 | \mu, \nu)$ obtained from the reverse process to approximate $q(x_0 | \mu, \nu)$. Since it is common that $p_{\theta} (x_0 | \mu, \nu)$ is computationally intractable, we follow the parameterization trick in~\cite{ho2020denoising, jeong2021diff} to calculate the variational lower bound of the log-likelihood. Specifically, the training objective of the diffusion model is L1 loss function between the noise $\epsilon \sim \mathcal{N}(0, I)$ and the diffusion model output $f_{\theta}$ described as follows:
\begin{align}
    \min_{\theta}|| \epsilon - f_{\theta}(h(x_0, \epsilon), t, \mu, \nu) ||_1,
\end{align}
where $h(x_0, \epsilon) = \sqrt{\hat{\beta_t}} x_0 + \sqrt{1 - \hat{\beta_t}}\epsilon$, and $\hat{\beta_t} = \prod^t_{\overline{t}=1} 1 - \beta_{\overline{t}}$.

\subsection{Training and Inference}
During training, we use physics priors extracted from the audio waveform as an additional condition to guide the model to learning correspondence between video inputs and impact sounds. However, since the ground truth sound clip is unavailable during inference, we could not obtain the corresponding physics priors for the video input as we did in the training stage. Therefore, we propose a new inference pipeline to allow us to preserve the benefit of physics priors. To achieve this goal, we construct key-value pairs for visual and physics latents in our training sets. At the inference stage, we feed the testing video input and acquire the visual latent vector $\nu^{\text{test}}$. We then take  $\nu^{\text{test}}$ as a query feature and find the key in training data by computing the Euclidean distance between the test video latent $\nu^{\text{test}}$ and all training video latents $\{\nu^{\text{train}}_j\}^J_{j=1}$. Given the key $\nu^{\text{train}}_j$, we then use the value $\mu^{\text{train}}_j$ as our test physics latent $\hat{\mu}^{\text{test}}$. Once we have both visual latent $\nu^{\text{test}}$ and physics latent $\hat{\mu}^{\text{test}}$, the model reverses the noisy spectrogram by first predicting the added noise at each forward iteration to get model output $f_{\theta} (x_t, t, \hat{\mu}^{\text{test}}, \nu^{\text{test}})$ and then removes the noise by the following:
\begin{align}
    x_{t-1} = \frac{1}{\sqrt{1 - \beta_t}}(x_t - \frac{\beta_t}{\sqrt{1 - \hat{\beta}_t}}f_{\theta}(x_t, t, \hat{\mu}^{\text{test}}, \nu^{\text{test}})) + \eta_t \epsilon_t,
\end{align}
where $\hat{\beta}_t = \prod_{\overline{t}=1}^t 1- \beta_{\overline{t}}$, $\epsilon_t \sim \mathcal{N}(0, I)$, $\eta = \sigma \sqrt{\frac{1-\hat{\beta}_{t-1}}{1-\hat{\beta}_t}\beta_t}$, and $\sigma$ is a temperature scaling factor of the variance~\cite{jeong2021diff}. After iterative sampling over all of the time steps, we can obtain the final spectrogram distribution $p_{\theta}(x_0 | \hat{\mu}^{\text{test}}, \nu^{\text{test}})$. It is worth noting that while we use the physics latent from the training set, we can still generate novel sound since the diffusion model also takes additional visual features as input.
\label{sec:method}

\section{Experiments}
\subsection{Dataset}
To evaluate our physics-driven diffusion models and make comparison to other approaches, we use the \textit{Greatest Hits} dataset~\cite{owens2016visually} in which people interact with physical objects by hitting and scratching materials with a drumstick, comprising 46,577 total actions in 977 videos. Human annotators labeled the actions with material labels and the time stamps of the impact sound. According to the dataset annotation assumption that the time between two consecutive object sounds is at least $0.25$ second, we segment all audios into $0.25$ second clips based on the data annotation for training and testing. We use the pre-defined train/test split for all experiments.

\subsection{Implementation Details}
We use Pytorch to implement all models in our method. For physics parameter estimation, all audio waveforms are in 44.1Khz and we compute log-scaled spectrograms with 2048 window size and 256 hop size, leading to a $1025 \times 44$ spectrogram for each impact sound. Then we estimate $1025$ modes parameters from the spectrogram as described in the Sec \ref{physics param estim}. For residual parameter prediction, the transformer encoder is a 4-layer transformer encoder with 4 attention heads. The residual weights and decay rate dimensions are both $100$. In the physics-driven diffusion model, we feed $22$ video frames centered at the impact event to the video encoder which is a ResNet-50 model with TSM~\cite{lin2019tsm} to efficiently handle the temporal information. The physics encoder consists of five parallel MLPs which take each of physics priors as input and project into lower-dimension feature vectors. The outputs are concatenated together into a $256$-dim physics latent vector $\mu$. The spectrogram denoiser is an Unet architecture, which is constructed as a spatial downsampling pass followed by a spatial upsampling pass with skip connections to the downsampling pass activation. We use Griffin-Lim algorithm to convert the spectrogram to the final audio waveform~\cite{griffin1984signal}. We use AdamW optimizer to train all models on a A6000 GPU with a batch size of 16 until convergence. The initial learning rate is set to $5e-4$, and it gradually decreases by a factor of 0.95.

\subsection{Baselines}
We compare our physics-driven diffusion model against various state-of-the-art systems. For fair comparison, we use the same video features extracted by TSM~\cite{lin2019tsm}.
\vspace{-2mm}
\begin{itemize}[align=right,itemindent=0em,labelsep=2pt,labelwidth=1em,leftmargin=*,itemsep=-0.5em]
    \item \textbf{ConvNet-based Model}: With a sequence video features, we first up-sampled them to have the same number of frames as the spectrogram. Then we perform a Unet architecture to convert video features to spectrogram. Such a architecture has shown successful results in spectrogram-based music generation~\cite{wang2019performancenet}.
    \item \textbf{Transformer-based Model}: We implement a conditional Transformer Network which has shown promising results in Text-to-Speech~\cite{li2019neural}. Instead of using text as condition, here we use the extracted video features.
    \item \textbf{Video conditioned Diffusion model}: We also compare our approach to two video conditioned spectrogram diffusion model variants. In the first setup, we do not include the physics priors and keep all other settings the same.
    \item \textbf{Video + Class Label conditioned Diffusion model}: In the second variant, we provide a class-label of the impact sound material as an additional condition to the video features. All other settings are the same as ours.
    \item \textbf{Video + Other Audio Features Diffusion model}: To show the importance of physics latents, we replace the physics latent with spectrogram/mfcc latent by extracting spectrogram/mfcc features from the raw audio and pass them to a transformer encoder similar to the one used in physics parameters reconstruction, and then we apply average pooling to obtain the latent vector. During testing, we still use visual features to query the corresponding spectrogram/mfcc latent in the training set and synthesize the final results.
\end{itemize}

\subsection{Evaluation Metrics}
We use four different metrics to automatically assess both the fidelity and relevance of the generated samples. For automatic evaluation purpose, we train an impact sound object material classifier using the labels in Great Hits Dataset. The classifier is a ResNet-50 convolutional-based neural network and we use the spectrogram as input to train the classifier.
\vspace{-2mm}
\begin{itemize}[align=right,itemindent=0em,labelsep=2pt,labelwidth=1em,leftmargin=*,itemsep=-0.5em] 
    \item \textbf{Fréchet Inception Distance (FID)} is used for evaluating the quality of generated impact sound spectrograms. The FID score evaluates the distance between the distribution of synthesized spectrograms and the spectrograms in the test set. To build the distribution, we extract the features before the impact sound classification layer.
    \item \textbf{Kernel Inception Distance (KID)} is calculated via maximum mean discrepancy (MMD). Again, we extract features from synthesized and real impact sounds. The MMD is calculated over a number of subsets to both get the mean and standard deviation of KID.
    \item \textbf{KL Divergence} is used to individually compute the distance between output distributions of synthesized and ground truth features since FID and KID mainly rely on the distribution of a collection of samples.
    \item \textbf{Recognition accuracy} is used to evaluate if the quality of generated impact sound samples can fool the classifier.
\end{itemize}

\subsection{Results}
\begin{figure*}[!t]
    \centering
    \includegraphics[width=0.8\textwidth]{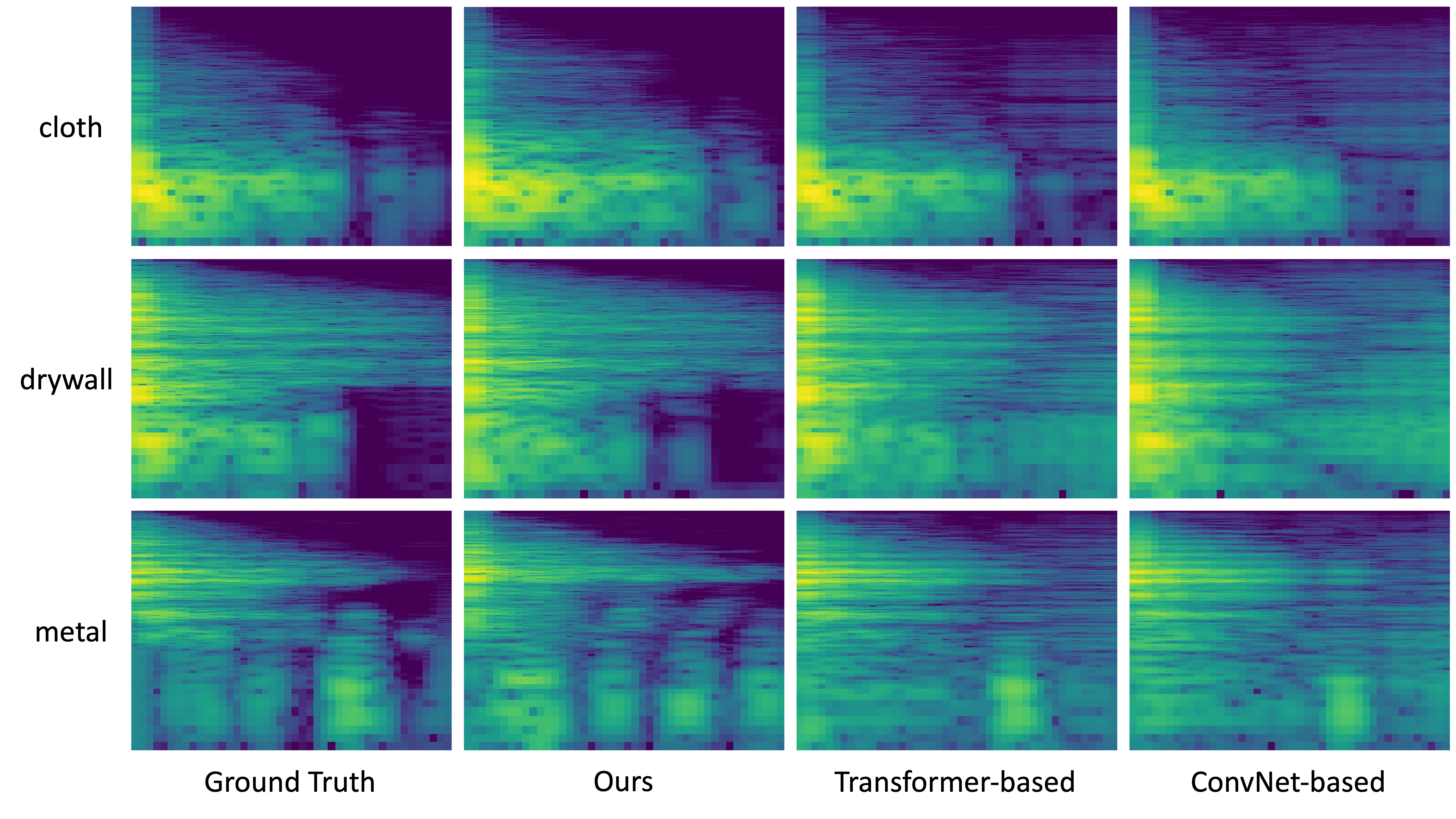}
    \vspace{-3mm}
    \caption{Qualitative Comparison results on sound spectrogram generated by different methods.}
    \vspace{-3mm}
    \label{fig:qualitative_comparison}
\end{figure*}

\begin{table*}[]
\centering
\small
{%
\begin{tabular}{|l|cccc|}
\hline
Model\textbackslash{}Metric & FID \textcolor{red}{$\downarrow$} & KID (mean, std)\textcolor{red}{$\downarrow$} & KL Div. \textcolor{red}{$\downarrow$} & Recog. Acc (\%) \textcolor{red}{$\uparrow$}    \\
\hline
ConvNet-based  & 43.50 & 0.053, 0.013 &  4.65 & 51.69    \\ 
Transformer-based  & 34.35 & 0.036, 0.015 & 3.13 & 62.86   \\
Video Diffusion & 54.57 & 0.054, 0.014 & 2.77 & 69.94 \\
Video + Class label Diffusion  & 31.82 & 0.026, 0.021 & 2.38 & 72.02  \\
Video + MFCC Diffusion & 40.21 & 0.037, 0.010 & 2.84 & 67.87 \\
Video + Spec Diffusion & 28.77 & 0.016, 0.009 & 2.55 & 70.46 \\
\cellcolor{mygray-bg}\bf{Video + Physics Diffusion~(Ours)}      & \cellcolor{mygray-bg}\bf 26.20  &   \cellcolor{mygray-bg}\bf 0.010, 0.008 & \cellcolor{mygray-bg}\bf 2.04 & \cellcolor{mygray-bg}\bf 74.09 \\ \hline
\end{tabular}%
}
\caption{Quantitative evaluations for different models. For FID, KID, and KL Divergence, lower is better. For recognition accuracy, higher is better. Bold font indicates the best value.}
\label{tab:quantitative results}
\end{table*}
Quantitative evaluation results are shown in Table~\ref{tab:quantitative results}. Our proposed physics-driven diffusion model outperforms all other methods across all metrics. It is worth noting that without physics priors, using video features alone as condition to the spectrogram denoiser is not sufficient to generate high fidelity sounds. While this could be improved when class labels are available, it is obvious that there is a large gap to reach the performance of the physics-driven method. Fig.~\ref{fig:qualitative_comparison} illustrates a comparison of three examples of generated spectrograms given a video by ConvNet-based, Transformer-based, and our physics-driven approaches to the ground truth. While the ConvNet and Transformer-based approaches could also capture some correspondences between audio and video, it is obvious that a lot of noise appears in the generated spectrogram because these approaches are prone to learn an average smoothed representation, and thus introducing many artifacts in the end. In comparison, our physics-driven diffusion approach does not suffer from this problem and can synthesize high fidelity sound from videos. It is worth noting that the interoperability of our approach could potentially unlock applications such as controllable sound synthesis by manipulating the physics priors.

\subsection{Human Perceptual Evaluation}
In addition to the objective evaluation of our method, we also perform human perceptual surveys using Amazon Mechanical Turk (AMT). We aim to use perceptual surveys to evaluate the effectiveness of our generated samples to match the video content and the fidelity of the generated samples. For all surveys, we do not request participants with any background on the survey or our approach was given to the participants to avoid perceptual biases. We surveyed $50$ participants individually, where each participant was asked to evaluate $10$ videos along with different generated samples from various methods. A total of $500$ opinions were collected in the end.
\vspace{-2mm}
\begin{itemize}[align=right,itemindent=0em,labelsep=2pt,labelwidth=1em,leftmargin=*,itemsep=-0.5em] 
    \item \textbf{Matching}. In the first survey, we asked people to watch the same video with different synthesized sounds and answer the question: "In which video the sound best matches the video content?". The participants would choose one soundtrack from the ConvNet-based, Transformer-based and Physics-driven diffusion approaches. From results shown in Table.~\ref{tab:human_eval_compare} (Top) (left column), we observe that there exists a clear indication that the sound generated with our method is selected as the best match to the visual content with a higher selected rate. 
    \item \textbf{Quality}. In the second survey, we asked people (non-expert) to choose the video with the highest sound quality, including 3 variations of samples generated by ConvNet-based, Transformer-based and Physics-driven diffusion approaches. Results in Table clearly indicate our approach achieves the best sound quality.
    \item \textbf{Perceptual Ablation Studies}. In the last survey, we performed a perceptual ablation study to test how the physics priors could influence the perception of the generated impact sounds compared to the approaches without it. Survey results shown in Table~\ref{tab:human_eval_compare} (Bottom) and suggest that in comparison to video only model, the physics priors improve the overall perception of synthesized impact sounds.
\end{itemize}

\begin{table}[]
\vspace{-3mm}
\centering
\footnotesize
\begin{tabular}{|l|c|c|}
\hline
Model\textbackslash{}Metric & Matching &  Quality \\
\hline
\multicolumn{3}{|l|}{\textit{Comparison to Baselines}}\\ \hline
ConvNet-based  & 18\% & 17.6\% \\
Transformer-based  & 26.6\% & 28.8\% \\ 
\cellcolor{mygray-bg}\bf{Ours}  & \cellcolor{mygray-bg}\bf 55.4\%  & \cellcolor{mygray-bg}\bf 53.6\% \\  \hline
\multicolumn{3}{|l|}{\textit{Perceptual Ablation Studies}}\\ \hline
Video-only  & 23.6\% & 23.6\% \\
 Video+label  & 37.8\% & 35.8\% \\ 
\cellcolor{mygray-bg}\bf{Ours}  & \cellcolor{mygray-bg}\bf 38.6\%  & \cellcolor{mygray-bg}\bf 40.6\% \\  \hline
\end{tabular}%
\vspace{-1mm}
\caption{(Top) Human perceptual evaluation on matching and quality metrics. (Bottom) Ablation study on human perceptual evaluation. The value indicates the percentage of Amazon Turkers who select the method.}
\vspace{-5mm}
\label{tab:human_eval_compare}

\end{table}


\subsection{Ablation Studies}
We performed three ablation studies to answer the following questions. \textbf{Q1}: How do residual parameters influence the physics priors? \textbf{Q2}: What is the contribution of each component to our approach? \textbf{Q3}: Is our method better than simple retrieval methods?

\begin{figure}[!t]
    \vspace{-3mm}
    \centering
    \includegraphics[width=\linewidth]{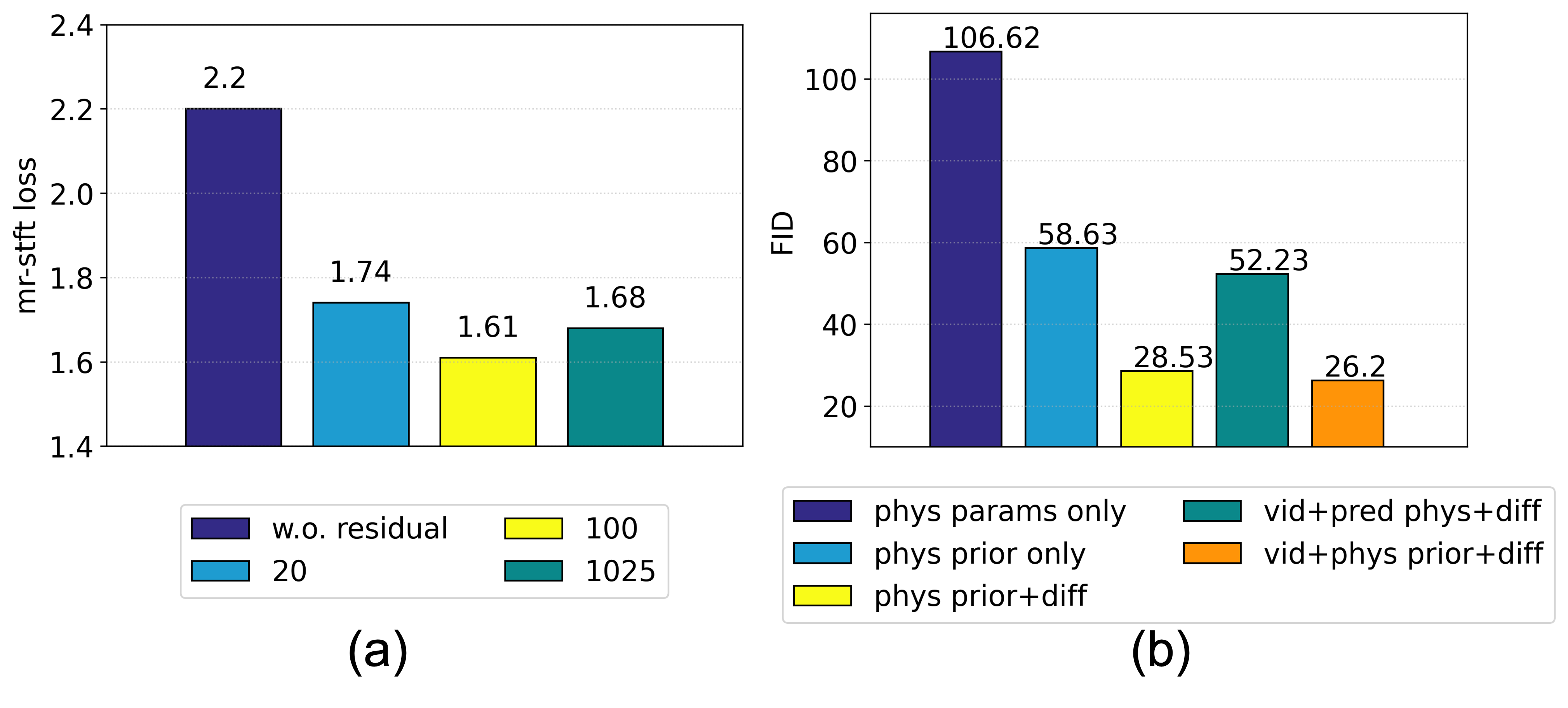}
    \vspace{-5mm}
    \caption{(a) Ablation study on the importance and selection for the number of residual parameters by testing multi-resolution STFT loss. (b) Ablation study on the contribution of each component of our approach using FID score, the lower the better.}
    \vspace{-5mm}
    \label{fig:ablation}
\end{figure}

\noindent\textbf{Q1}. Since the physics and residual parameters are essential in our approach. We have investigated different physics priors variants to find the most informative physics priors and use the multi-resolution STFT loss of the reconstructed sounds for evaluation. The results in Fig.~\ref{fig:ablation}(a) clearly show that the loss decreases significantly with residual parameters. We also find that using $100$ residual parameters achieves the best performance, while fewer or more residual parameters may damage the performance.

\noindent\textbf{Q2}. We perform extensive experiments to understand the contribution of each component. For all studies, we use the nearest physics parameters/priors retrieved by visual latent to synthesize the sound. Results are shown in Fig.~\ref{fig:ablation}(b). We first observe that without residual components and diffusion models, using estimated physics parameters to perform modal synthesis could not obtain faithful impact sounds. The physics priors could re-synthesize impact sounds with much better quality with learned residual parameters. We further show that using physics priors as the condition input to the diffusion model achieves even better performance. We have also performed an experiment to predict physics latent from video input and use it as the condition for the diffusion model but the quality of generated samples is poor. This is due to the weak correspondence between video inputs and physics behind the impact sounds and indicates the importance of using video inputs to query physics priors of the training set at the inference stage.

\noindent\textbf{Q3}. We consider two retrieval baselines for comparison. The first one is a simple baseline without physics priors and diffusion model. We only use visual features extracted from the ResNet-50 backbone to search the nearest neighbor (NN) in the training set and use the audio as output. In the second experiment, we try our best to reproduce the model in~\cite{owens2016visually} since no official implementation is available. The model predicts sound features (cochleagrams) from images via LSTM. For fair evaluation, a collection-based metric like FID is invalid because the retrieved audios are in the real data distribution. Therefore, we use sample-based metrics, including KL Divergence between predicted and ground truth audio features and Mean Square Error on the spectrogram level. The table~\ref{tab:retrieval_compare} clearly shows that our approach outperforms the retrieval baselines by a large margin.
\begin{table}[h]
\vspace{-3mm}
\centering
\footnotesize
{%
\begin{tabular}{|l|cc|}
\hline
Model\textbackslash{}Metric & KL Div. \textcolor{red}{$\downarrow$} &  Spec. MSE\textcolor{red}{$\downarrow$} \\
\hline
NN via Visual Features & 10.60 & 0.307 \\
NN via Predicted Sound Features~\cite{owens2016visually} & 7.39 & 0.205 \\
\cellcolor{mygray-bg}\bf{Ours}  & \cellcolor{mygray-bg}\bf 2.04  & \cellcolor{mygray-bg}\bf 0.149 \\  \hline
\end{tabular}%
}
\vspace{-3mm}
\caption{Comparison with retrieval methods.}
\label{tab:retrieval_compare}
\vspace{-4mm}
\end{table}

\section{Conclusion}
We present a physics-driven diffusion model for impact sound synthesis from videos. Our model can effectively generate high fidelity sounds for physical object interactions. We achieve such function by leveraging physics priors as guidance for the diffusion model to generate impact sounds from video input. Experimental results demonstrate that our approach outperforms other methods quantitatively and qualitatively. Ablation studies have demonstrated that physics priors are critical for generating high-fidelity sounds from video inputs. The limitation of our approach naturally becomes that our approach cannot generate impact sounds for unseen physics parameters due to the query process (failure case demonstration is shown in Supplementary material), while we can generate novel sounds given an unseen video.
\label{sec:conclusion}
\paragraph{Acknowledgements.} This work was supported by the MIT-IBM Watson AI Lab, DARPA MCS, DSO grant DSOCO21072, and gift funding from MERL, Cisco, Sony, and Amazon. 

{\small
\bibliographystyle{ieee_fullname}
\bibliography{main}
}

\clearpage 

\appendix
\label{sec:appendix}
\section{Generated Samples}
Please see the \textbf{attached video for a short overview and generated impact sound samples using our physics-driven diffusion models.} Please turn \textbf{Audio ON} and use \textbf{headphones} for best perception of the audio.

\section{Fixed Physics Priors}
As we mentioned in the main paper, while we use the physics latent from training set, we can still generate novel sound since the diffusion model also takes additional visual features as input. To show that, we fix one type physics priors and use various video inputs to synthesize impact sounds. We found that the synthesized sounds are strongly impacted by the physics priors but different visual inputs can still introduce the diversity of the sounds. Please check out the videos with prefix \textbf{fix\_physics\_priors\_type\_sample\_num.mp4}.

\section{Physics Priors Editing}
\begin{figure}[h]
    \centering
    \includegraphics[width=0.8\linewidth]{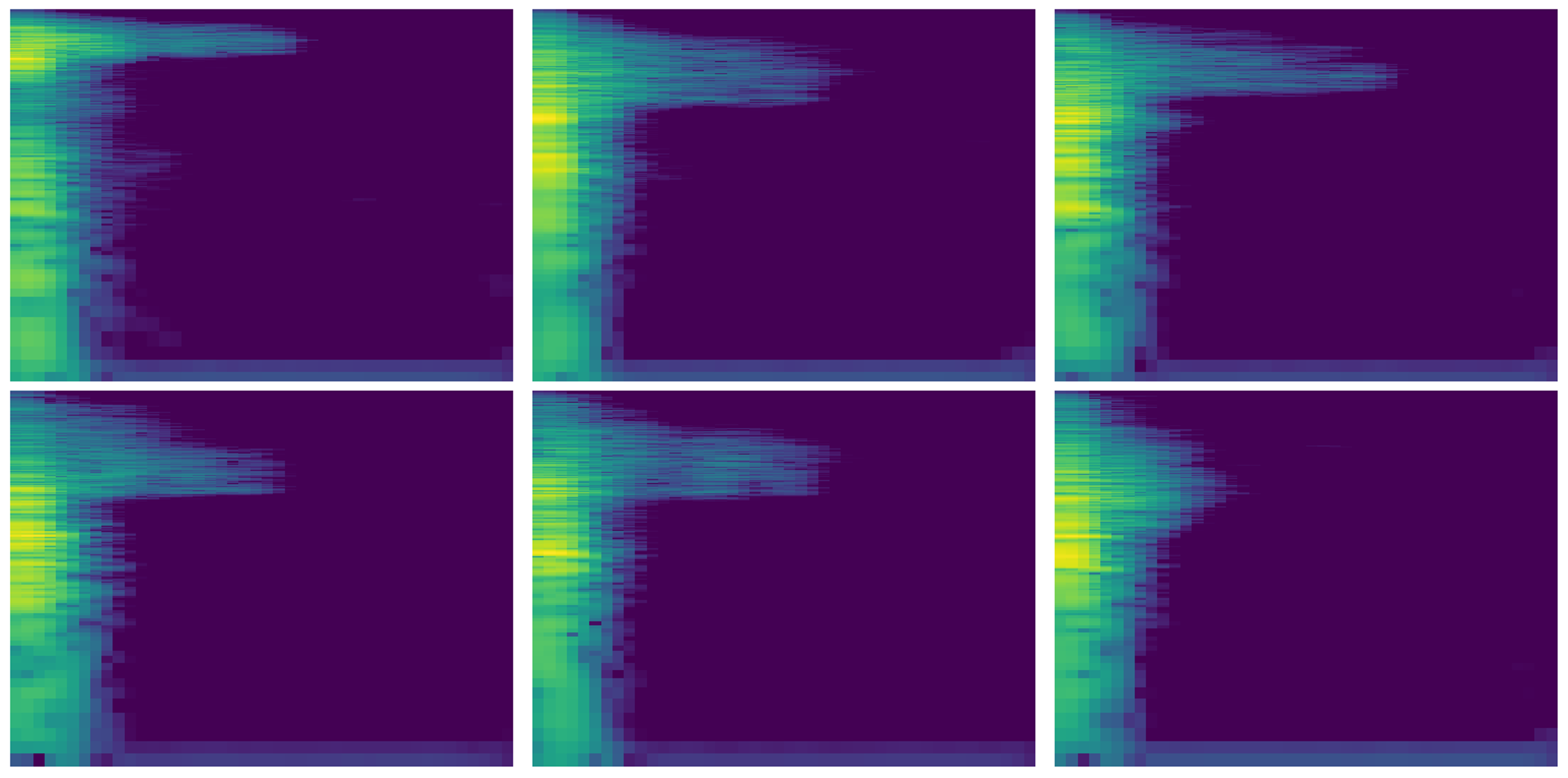}
    \caption{Spectrogram examples of removing low frequency parts by manipulating physics priors.}
    \label{fig:remove_low_freq}
\end{figure}

\begin{figure}[h]
    \centering
    \includegraphics[width=0.8\linewidth]{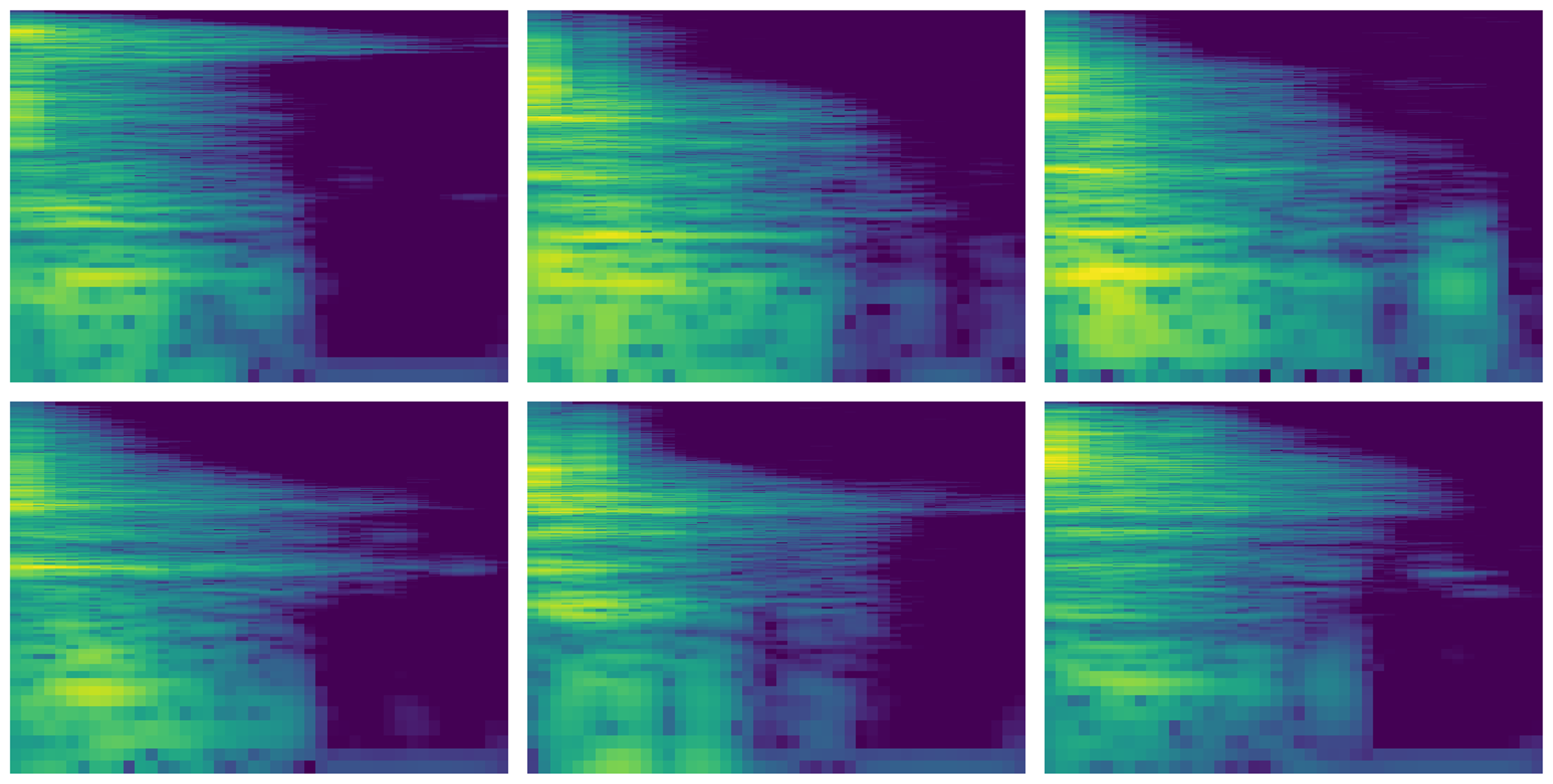}
    \caption{Spectrogram examples of adding low frequency parts by manipulating physics priors.}
    \label{fig:add_low_freq}
\end{figure}

Since the physics priors are transparent, we can manipulate them to control the synthesized sounds. In the first experiment, we decrease values of power and decay rate in the lowest $200$ modes to reduce low frequency components and we also set the residual parameters be zero for better visualization. The generated spectrogram results are shown in Fig.~\ref{fig:remove_low_freq}. In addition, we also try highlighting the low frequency components by tuning up the power and decay values of the lowest $200$ modes. The generated spectrogram results are shown in Fig.~\ref{fig:add_low_freq}. Both results demonstrate that the transparency of the physics priors.
\section{Additional Implementation Details}
\subsection{Physics Priors Representation \& Encoder}
\begin{figure}[h]
    \centering
    \includegraphics[width=0.6\linewidth]{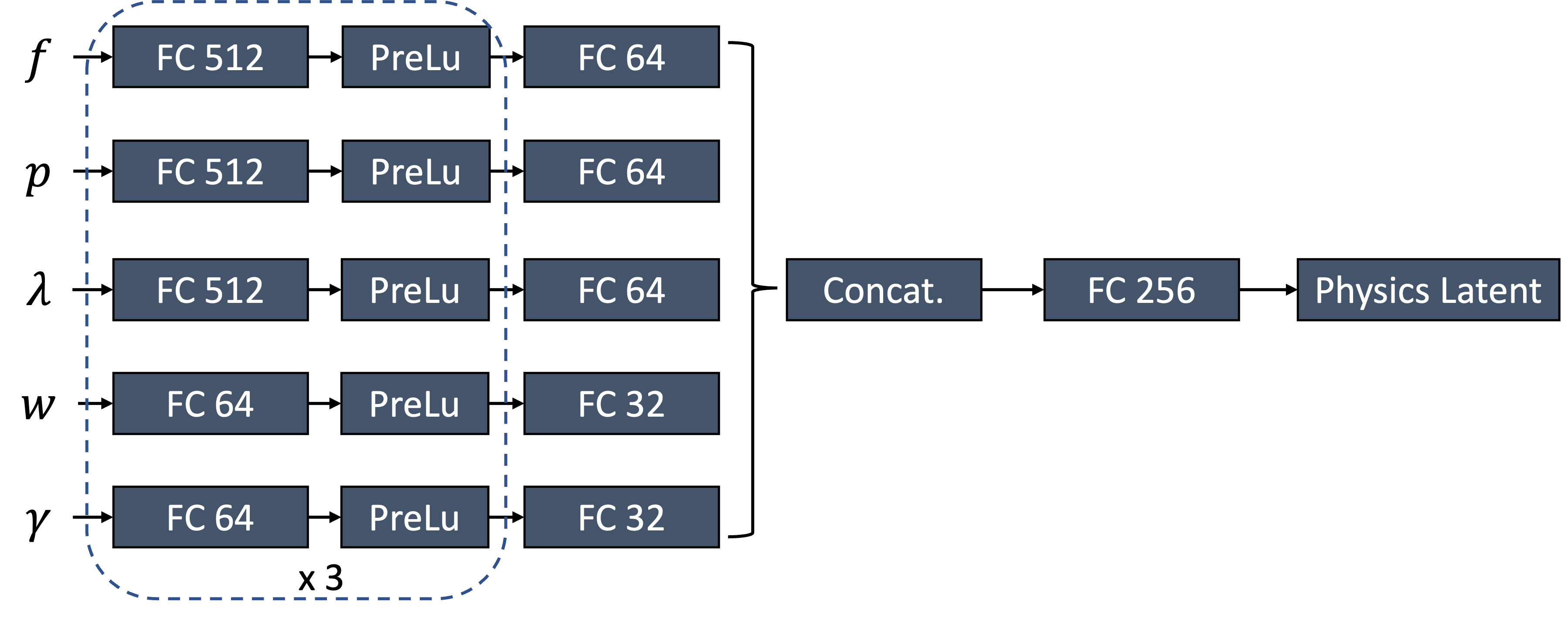}
    \caption{Physics Priors Encoder Details}
    \label{fig:physics_priors_enc}
\end{figure}

We provide more details of how physics parameters and residual parameters are scaled and fitted into the diffusion model. For each estimated frequency feature $f$, we compute the distance to the corresponding central frequency of the spectrogram bin and then normalize the distance by the bin resolution (21Hz in our case) to obtain a value between $-1$ and $1$. The power feature $p$ is extracted in dB scale. We normalize it by $2p/(-80) - 1$ where $-80$dB indicates the silence. The decay feature $\lambda$ is also applied min-max normalization to map the value to the range of $-1$ and $1$. For residual parameters, we use a transformer spectrogram encoder to learn $w$ and $\gamma$ by applying a sigmoid function to outputs of two linear projection layers. We then use a physics priors encoder to take produce the physics latent. The details of the physics priors encoder are shown in Fig.~\ref{fig:physics_priors_enc}.

\subsection{Diffusion Model Details}
The architecture is a U-Net~\cite{ronneberger2015u} based on a Wide ResNet~\cite{zagoruyko2016wide}. Our models use four feature map resolutions and each resolution level contains two convolutional residual blocks and a spatial attention module. The channel dimensions increase from $1$ to $64$, $128$, $256$, and $512$. We use sinusoidal position embedding for diffusion time $t$ which is added into each residual block. We use cosine schedule for diffusion noise and set the sampling timesteps be $T=1000$.
\section{Failure case demonstration} We show an example of having an impact on rock. In the following figure, the (a, b) are common rock sounds in the training set. However, we find a rare case in the testing set shown in (c). While our approach can generate new realistic rock sound (d), it is different from the ground truth. 
\begin{figure}[h]
    \centering
    \includegraphics[width=0.85\linewidth]{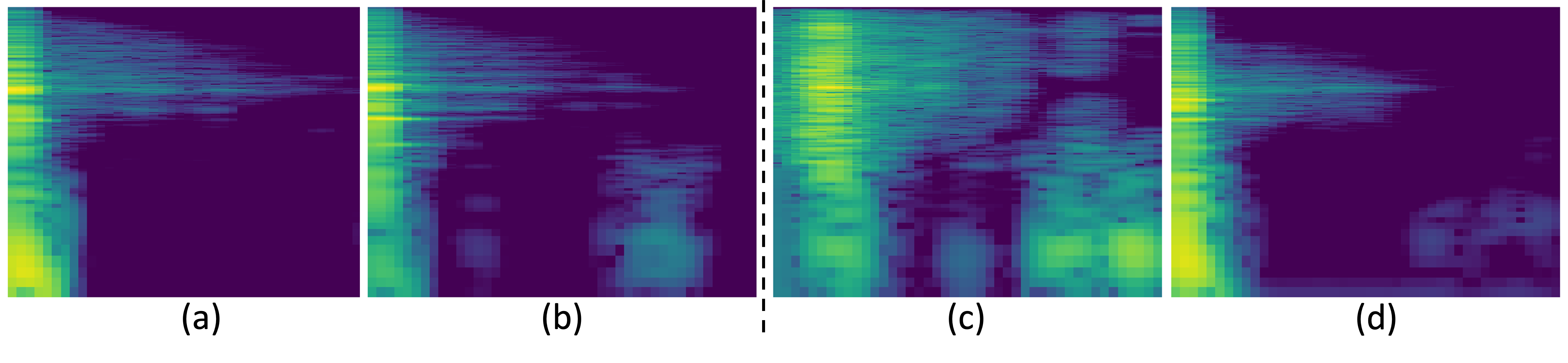}
    \label{fig:failure}
\end{figure}


\end{document}